\pdfoutput=1

\documentclass[11pt]{article}

\usepackage{naacl2021}

\usepackage{times}
\usepackage{latexsym}

\usepackage[T1]{fontenc}

\usepackage[utf8]{inputenc}

\usepackage{microtype}

\usepackage{url}
\usepackage{graphicx}
\usepackage{wrapfig}
\usepackage{amsmath}
\usepackage{amssymb}
\usepackage{color,xcolor,colortbl}
\usepackage{enumitem}
\usepackage{algorithm}
\usepackage{algorithmic}
\usepackage{bm,bbm}
\usepackage{booktabs}
\usepackage{mathtools}
\usepackage{array}
\usepackage{multirow}
\usepackage{soul}
\usepackage{subcaption}
\usepackage{pbox}
\usepackage{pifont}
\usepackage{arydshln}

\definecolor{darkblue}{rgb}{0.0, 0.0, 0.55}
\newenvironment{fontppl}{\fontfamily{ppl}\selectfont}{\par} 
\setul{0.5ex}{0.3ex} 
\setulcolor{blue} 
\definecolor{darkblue}{rgb}{0.0, 0.0, 0.55}
\definecolor{midnightblue}{HTML}{191970}
\definecolor{darkgreen}{HTML}{006400}
\definecolor{red}{HTML}{ff0000}
\definecolor{gold}{HTML}{ffd700}
\definecolor{mediumvioletred}{HTML}{c71585}
\definecolor{lime}{HTML}{00ff00}
\definecolor{aqua}{HTML}{00ffff}
\definecolor{fuchsia}{HTML}{ff00ff}
\definecolor{lightpink}{HTML}{ffb6c1}
\definecolor{dodgerblue}{HTML}{1e90ff}
\definecolor{deepskyblue}{HTML}{00bfff}
\definecolor{deeppink}{HTML}{ff1493}
\definecolor{orangered}{HTML}{ff4500}
\definecolor{mediumseagreen}{HTML}{3cb371}
\definecolor{saddlebrown}{HTML}{8b4513}

\usepackage[T1]{fontenc}
\usepackage[scaled=.8]{beramono}

\title{A Sliding-Window Approach to Automatic Creation of Meeting Minutes}

\author{Jia Jin Koay, Alexander Roustai, Xiaojin Dai, Fei Liu\\[0.5em]
Computer Science Department\\
University of Central Florida,
Orlando, FL 32816\\[0.5em]
\texttt{\{jjkoay,alexroustai,xd.zangyiwu\}@knights.ucf.edu}\\
\texttt{feiliu@cs.ucf.edu}
}

\begin{document}
\maketitle
\begin{abstract}

Meeting minutes record any subject matters discussed, decisions reached and actions taken at meetings.
The importance of minuting cannot be overemphasized in a time when a significant number of meetings take place in the virtual space.
In this paper, we present a sliding window approach to automatic generation of meeting minutes.
It aims to tackle issues associated with the nature of spoken text, including lengthy transcripts and lack of document structure, which make it difficult to identify salient content to be included in the meeting minutes.
Our approach combines a sliding window and a neural abstractive summarizer to navigate through the transcripts to find salient content. 
The approach is evaluated on transcripts of natural meeting conversations, where we compare results obtained for human transcripts and two versions of automatic transcripts and discuss how and to what extent the summarizer succeeds at capturing salient content.

\end{abstract}

\section{Introduction}

Meetings are ubiquitous across organizations of all shapes and sizes, and it takes a tremendous effort to record any subject matters discussed, final decisions reached and actions taken at meetings.
With the rise of remote workforce, virtual meetings are more important than ever. 
An increasing number of video conferencing providers including Zoom, Microsoft Team, Amazon Chime and Google Meet allow meetings to be transcribed~\cite{Martindale-2021}. 
However, without automatic minuting, consolidating notes and creating meeting minutes is still regarded as a tedious and time-consuming task for meeting participants.
There is thus an urgent need to develop advanced techniques to better summarize and organize meeting content.

Meeting summarization has been attempted on a small scale before the era of deep learning. 
Previous work includes efforts to 
extract utterances and keyphrases from meeting transcripts~\cite{galley-2006-skip,murray-carenini-2008-summarizing,gillick2009,liu-etal-2009-unsupervised},
detect meeting decisions~\cite{Hsueh-2008},
compress or merge utterances to generate abstracts~\cite{liu-liu-2009-extractive,wang-cardie-2013-domain,mehdad-etal-2013-abstractive}
and make use of acoustic-prosodic and speaker features~\cite{Maskey05comparinglexical,zhu-etal-2009-summarizing,DBLP:conf/slt/ChenM12} for utterance extraction.
The continued development of automatic transcription and its easy accessibility have sparked a renewed interest in meeting summarization~\cite{shang-etal-2018-unsupervised,li-etal-2019-keep,koay-etal-2020-domain,song-etal-2020-summarizing,zhu-etal-2020-hierarchical,zhong2021}, where neural representations are explored for this task.
We believe the time is therefore ripe for a reconsideration of the approach to automatic minuting.

It may be tempting to apply neural abstractive summarization to meetings given its remarkable recent success on summarization benchmarks, e.g., CNN/DM~\cite{see-etal-2017-get,chen-bansal-2018-fast,gehrmann-etal-2018-bottom,laban-etal-2020-summary}.
However, the challenge lies not only in handling hallucinations that are seen in abstractive models~\cite{kryscinski-etal-2019-neural,lebanoff-etal-2019-analyzing,maynez-etal-2020-faithfulness} but also the models' strong positional bias that occurs as a consequence of fine-tuning on news articles~\cite{kedzie-etal-2018-content,grenander-etal-2019-countering}.
Neural summarizers also assume a maximum sequence length, e.g., Perez-Beltrachini et al.~\shortcite{perez-beltrachini-etal-2019-generating} use the first 800 tokens of the document as input.
With an estimated speaking rate of 122 words per minute~\cite{polifroni-etal-1991-collection}, it indicates that the summarizer may only process a relatively short transcript 
-- about 5 minutes in duration.

In this paper, we instead study an extractive meeting summarizer to identify salient utterances from the transcripts.
It leverages a sliding window to navigate through a transcript of any length and a neural abstractive summarizer to find salient local content.
In particular, we aim to address three key questions:
(1) what are suitable window and stride sizes?
(2) can the abstractive summarizer effectively identify salient local content? 
(3) how should we consolidate local abstracts into meeting-level summaries?
Our approach is intuitive and appealing, as humans make a sequence of local decisions when navigating through very long recordings.
It is evaluated on transcripts of natural meeting conversations~\cite{1198793}, where we obtained human transcripts and two versions of automatic transcripts produced by the AMI speech recognizer~\cite{10.1007/11677482_38} and Google Cloud's Speech-to-Text API.\footnote{\url{https://cloud.google.com/speech-to-text}}
Our contributions in this paper are as follows.

\begin{itemize}[topsep=5pt,itemsep=0pt,leftmargin=*]

\item We study the feasibility of a sliding-window approach to automatic generation of meeting minutes that draws on a pretrained neural abstractive summarizer to make local decisions on utterance saliency.
It does not require any annotated data and can be extended to meetings of various types and domains.

\item We examine results obtained from human transcripts and two versions of automatic transcripts, and show that
our summarizer either outperforms or performs comparably to competitive baselines given both automatic and human evaluations.
We discuss how and to what extent the summarizer succeeds at capturing salient content.\footnote{
Our transcripts and system outputs are released publicly at \url{https://github.com/ucfnlp/meeting-sliding-window}}

\end{itemize}

\section{Background: The BART Summarizer}

BART~\cite{lewis-etal-2020-bart} has demonstrated strong performance on neural abstractive summarization.
It consists of a bidirectional encoder and a left-to-right autoregressive decoder, each contains multiple layers of Transformers~\cite{NIPS2017_3f5ee243}.
The model is pretrained using a denoising objective that, given a corrupted input text, the encoder strives to learn meaningful representations and the decoder reconstructs the original text using the representations. 
In this study, we use \texttt{BART-large-cnn} as a base summarizer. 
It contains 12 layers in each of the encoder and decoder and uses a hidden size of 1024. The model is then fine-tuned on the CNN dataset for abstractive summarization.

There are two obstacles that should be overcome in order for BART to generate meeting summaries from transcripts.
Firstly, BART is trained on written text, rather than spoken text. 
The pretraining data contain 160G of news, books, stories, and web text.
It remains unclear if the model can effectively identify salient content on spoken text and, how it is to reduce lead bias that is not as frequent in spoken text as in news writing~\cite{grenander-etal-2019-countering}.
Secondly, a transcript can far exceed the maximum input length of the model, which is restricted by the GPU memory size.
This is the case even for recent variants such as Reformer~\cite{Kitaev2020Reformer} and Longformer~\cite{beltagy2020longformer}.

\begin{figure}[t]
\centering
\includegraphics[width=2.5in]{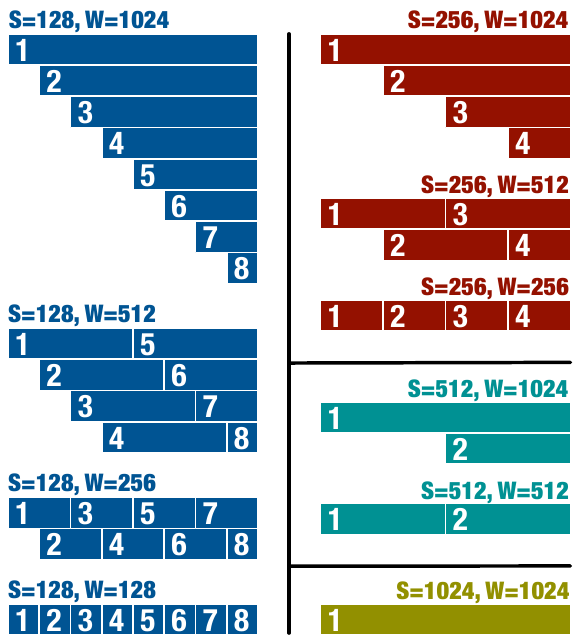}
\caption{A total of 10 combinations of window (\texttt{W}) and stride (\texttt{S}) sizes examined in this study.
A small stride allows a text region to be repeatedly visited by the summarizer.
The numbers (\texttt{1-8}) indicate local windows.
}
\label{fig:settings}
\vspace{-0.1in}
\end{figure}

\begin{table*}[t]
\setlength{\tabcolsep}{6.5pt}
\renewcommand{\arraystretch}{1.1}
\centering
\begin{fontppl}
\begin{footnotesize}
\begin{tabular}{|l|l|ccc|ccc|rr|}
\hline
& & \multicolumn{3}{c|}{\textbf{ROUGE-1}} & \multicolumn{3}{c|}{\textbf{ROUGE-2}} & \multicolumn{2}{c|}{\textbf{Summary Len}}\\
Input & System & P(\%) & R(\%) & F(\%) & P(\%) & R(\%) & F(\%) & \textbf{\%Uttrs} & \textbf{\#Wrds}\\
\hline
\hline
\multirow{7}{*}{Human} & KL-Sum & 57.2 & 31.9 & 40.8 & 19.0 & 10.6 & 13.6 & 19.6 & 754\\
& SumBasic & 61.6 & 67.1 & 62.4 & 24.8 & 28.1 & 25.6 & 19.6 & 1,730\\
& LexRank & 36.8 & 84.3 & 50.9 & 21.2 & 49.2 & 29.4 & 19.6 & 3,528\\
& TextRank & 28.2 & 91.6 & 42.9 & 19.4 & 63.5 & 29.5 & 19.6 & 4,954\\
& \cite{koay-etal-2020-domain} & 52.6 & 81.0 & 62.5 & 29.4 & 46.1 & 35.2 & 21.7 & 2,321\\ 
\cdashline{2-10}
& \textbf{SW (HumanTrans)} & \textbf{36.5} & \textbf{90.9} & \textbf{51.9} & \textbf{23.2} & \textbf{58.4} & \textbf{33.1} & 19.6 & 3,741\\
\hline
\hline
\multirow{4}{*}{ASR} & \cite{shang-etal-2018-unsupervised} & 27.6 & 36.3 & 31.0 & 4.4 & 5.6 & 4.8 & n/a & n/a\\
& \cite{koay-etal-2020-domain} & 51.3 & 78.6 & 61.3 & 25.7 & 39.9 & 30.9 & 16.7 & 2,224\\
\cdashline{2-10}
& \textbf{SW (AMI ASR)} & \textbf{36.1} & \textbf{88.3} & \textbf{51.2} & \textbf{19.4} & \textbf{47.8} & \textbf{27.6} & 18.2 & 3,514\\ 
& \textbf{SW (Google ASR)} & \textbf{61.9} & \textbf{65.7} & \textbf{62.9} & \textbf{26.5} & \textbf{28.1} & \textbf{26.9} & 23.2 & 1,460\\ 
\hline
\end{tabular}
\end{footnotesize}
\end{fontppl}
\vspace{-0.05in}
\caption{
Results on the ICSI test set using human transcripts and two versions of automatic transcripts (AMI vs. Google) as input.
The length is defined as percentage of selected utterances over all utterances of the meetings and average number of words in the summaries.
The sliding-window (SW) summarizer uses (\texttt{S}=128, \texttt{W}=1024).
}
\label{tab:results_test} 
\end{table*}

\section{Our Approach}

A sliding-window approach to generating meeting minutes is appealing because it breaks lengthy transcripts into small and manageable local windows, allowing a set of ``mini-summaries'' to be produced from such windows which are then assembled into meeting-level summaries.
There are two essential decisions to be made when using a sliding window.
Firstly, one must decide on the size of the local window. 
Our window size is bounded by the maximum sequence length of BART as the utterances in a window are concatenated into a flat sequence that serves as input to it.
We consider a number of window sizes with \texttt{W}=\{128, 256, 512, 1024\} tokens.
Secondly, a transcript may be partitioned into non-overlapping or partially overlapping windows.
We set the stride size to be \texttt{S}=\{128, 256, 512, 1024\} tokens to support both (\texttt{W} $\ge$ \texttt{S}). 
When they are of equal size, a transcript is divided into a sequence of non-overlapping windows.

In Figure~\ref{fig:settings}, we enumerate all 10 combinations of window and stride sizes. 
For example, we experiment with four window sizes of 128, 256, 512 and 1,024 tokens using the same stride size of 128 tokens, shown in dark blue (left).
A larger window gives additional context to BART for recognizing salient content. 
Using a window of 1,024 and stride of 128 tokens allow each utterance of the transcript to be visited 8 times, whereas using a window of 512 tokens reduces that to 4 times.

\vspace{0.08in}
\noindent\textbf{Consolidation.}\quad
BART abstracts generated from local windows cannot be simply concatenated to form meeting-level summaries as they contain redundancy. 
When local windows are partially overlapping, they can cause the same content to be included in different abstracts. 
Instead, we identify \emph{supporting utterances} of each abstract from the transcript. 
Particularly, we compute the ROUGE-L scores between each utterance in the window and the abstract. 
If the utterance is longer than 5 tokens, achieves a recall score $r$ > 0.5 and precision score $p$ > 0.1, we call it a supporting utterance.\footnote{
The thresholds were determined heuristically on the training set by observing the resulting alignment.}
The same utterance can support multiple abstracts.
We include an utterance into the meeting summary if it is designated as the supporting utterance for at lease one local abstract.
It lends flexibility and improves ease of consolidation of local abstractive summaries produced by BART.

\begin{figure*}[t]
\centering
\includegraphics[width=4.5in]{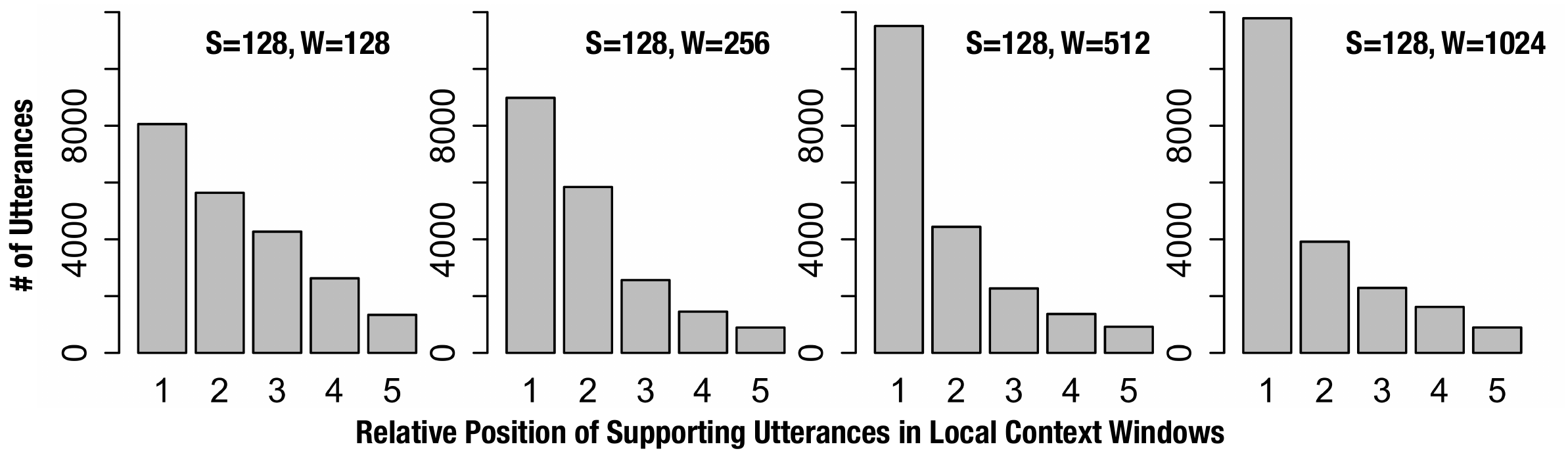}\\
\vspace{0.1in}
\includegraphics[width=5.5in]{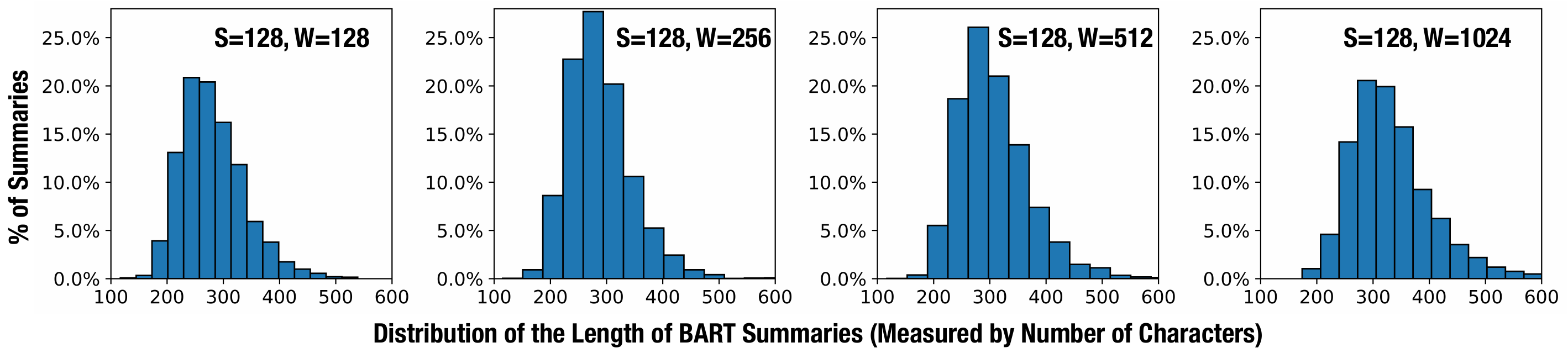}
\vspace{-0.05in}
\caption{(\textsc{Top}) Relative position of supporting utterances in their local windows.
We find that BART tends to take summary content from the first 150-200 tokens of the input sequence. 
With a large window (\texttt{W}=1024), summary content is likely taken from the first 20\% of input. 
(\textsc{Bottom}) Length distribution of BART abstracts, measured by number of characters. 
Using windows ranging from 128 to 1024 tokens, the average abstract length increases from 281 to 332 characters,
i.e., 56 to 66 words assuming 5 characters per word for English texts (Shannon, 1951).
Results are obtained on the ICSI training set using human transcripts.
}
\label{fig:len_pos}
\vspace{-0.1in}
\end{figure*}

\section{Results}

\noindent\textbf{Dataset.}
Our experiments are performed on the ICSI meeting corpus~\cite{1198793}, which is a challenging benchmark for meeting summarization.
The corpus contains 75 meeting recordings, each is about an hour long.
We use 54 meetings for training and report results on the standard test set containing 6 meetings.
Each training meeting has been annotated with an extractive summary.
Each test meeting has three human-annotated extractive summaries, which we use as gold-standard summaries.
The original corpus include human transcripts and automatic speech recognition (ASR) output generated by the AMI ASR team~\cite{10.1007/11677482_38}.
We are able to generate a new version of automatic transcripts by using Google's Speech-to-Text API as an off-the-shelf system.\footnote{
Due to lack of documentation, we are unable to report the word error rates of Google and AMI speech recognizers.}
Comparing results on different versions of transcripts allows us to better assess the generality of our findings.

Our baselines include both general-purpose extractive summarizers and meeting-specific summarizers.  
LexRank~\cite{Erkan:2004} and TextRank~\cite{mihalcea-tarau-2004-textrank} are graph-based extractive methods.
SumBasic~\cite{Vanderwende:2007} selects sentences if they contain frequently occurring content words.
KL-Sum~\cite{Haghighi:2009} adds sentences to the summary to minimize KL divergence.
We additionally experiment with two meeting summarizers.
Shang et al.~\shortcite{shang-etal-2018-unsupervised} group utterances into clusters, generate an abstractive sentence from each cluster using sentence compression, then select best elements from these sentences under a budget constraint. 
Koay et al.~\shortcite{koay-etal-2020-domain} develop a supervised BERT summarizer to identify summary utterances.

We report test set results in Table~\ref{tab:results_test}, where system summaries are compared with gold-standard extractive summaries using ROUGE metrics~\cite{lin-2004-rouge}.
The summary length is computed as the percentage of selected utterances over all utterances of the meetings and average number of words per test summary.
This information is reported wherever available, 
and baseline summarizers are set to output the same number of summary utterances as the sliding-window (SW) approach.
Our SW approach can outperform or perform comparably to competitive baselines when evaluated on human and ASR transcripts. 
We note that Koay et al.~\shortcite{koay-etal-2020-domain} utilize a supervised BERT summarizer, whereas our SW approach is unsupervised.\footnote{
We use pyrouge with default options to evaluate all summaries. The scores are different from that of Koay et al.~\shortcite{koay-etal-2020-domain} which removed stopwords during evaluation by using `-s'.}
It does not require annotated summaries and only uses the training set to determine window and stride sizes (\texttt{S}=128, \texttt{W}=1024, details later).

A closer examination reveals that Google transcripts contain substantially less filled pauses (\emph{um, uh, mm-hmm}), disfluencies (\emph{go-go-go away}), repetitions and verbal interruptions.
The Google service also tends to produce lengthier utterances.
Table~\ref{tab:stat_trans} provides an example comparing human, AMI and Google transcripts.
The summaries produced with Google transcripts contain fewer utterances and less number of words per summary. 
They achieve a higher precision and lower recall when compared to those of AMI and human transcripts.

\begin{table}[t]
\setlength{\tabcolsep}{3pt}
\renewcommand{\arraystretch}{1.05}
\centering
\begin{fontppl}
\begin{footnotesize}
\begin{tabular}{|l|ccc|}
\hline
\textbf{Transcription} & \textbf{Human} & \,\,\,\,\,\textbf{AMI}\,\,\, & \textbf{Google} \\
\hline
\hline
\# of utter. per meeting & 1330 & 1410 & 188\\
\# of words per utterance & 7.7 & 7.0 & 33.0 \\
\hline
\multicolumn{4}{|l|}{(\textbf{Human}) and um}\\
\multicolumn{4}{|l|}{There one of our}\\
\multicolumn{4}{|l|}{diligent workers has to sort of volunteer to}\\
\multicolumn{4}{|l|}{look over Tilman's shoulder while he is changing}\\
\multicolumn{4}{|l|}{the grammars to English}\\
\hline
\hline
\multicolumn{4}{|l|}{(\textbf{AMI}) And um}\\
\multicolumn{4}{|l|}{And they're one of our a}\\
\multicolumn{4}{|l|}{The legend to work paris has to sort of volunteer to}\\
\multicolumn{4}{|l|}{Look over time and shorter what he is changing}\\
\multicolumn{4}{|l|}{that gram was to english}\\
\hline
\hline
\multicolumn{4}{|l|}{(\textbf{Google}) and they are one of our diligent workers has}\\
\multicolumn{4}{|l|}{to sit or volunteer to look over two months shoulder}\\
\multicolumn{4}{|l|}{while he is changing the Grandma's to English}\\
\hline
\end{tabular}
\end{footnotesize}
\end{fontppl}
\vspace{-0.05in}
\caption{Compared to human and AMI transcripts, utterances produced by Google's transcription service are lengthier and there are fewer utterances per meeting.
}
\label{tab:stat_trans}
\vspace{-0.1in}
\end{table}

\begin{figure}[t]
\centering
\includegraphics[width=3in]{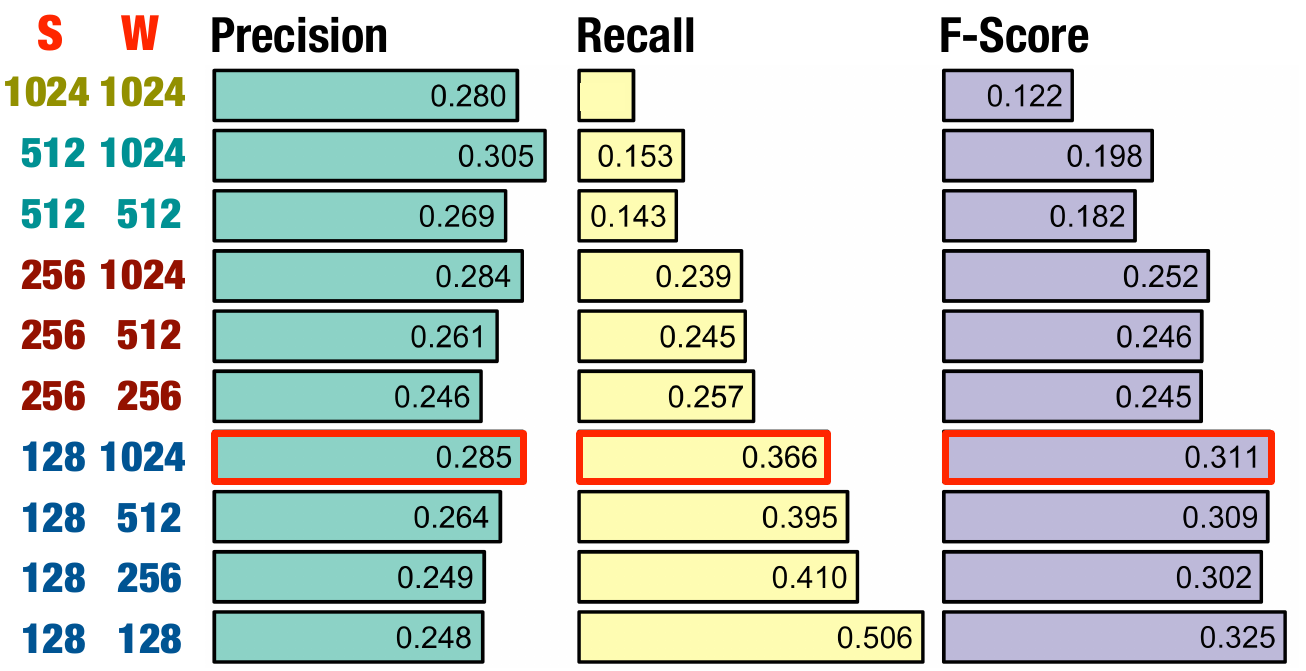}
\vspace{-0.05in}
\caption{Precision, recall and F-scores of summary utterance selection using different combinations of stride (\texttt{S}) and window (\texttt{W}) sizes.
Results are obtained on the ICSI training set using human transcripts.
We find that (\texttt{S}=128, \texttt{W}=1024) attains a good balance between precision and recall, whereas using small, non-overlapping windows (\texttt{S}=128, \texttt{W}=128) yields high recall due to more utterances are included in the summary.
}
\label{fig:results_cls_train}
\vspace{-0.1in}
\end{figure}

We are curious to know where supporting utterances appear in the local windows.
In Figure~\ref{fig:len_pos}, we discretize the position information into 5 bins and plot the distributions for four settings that use different window sizes (\texttt{W}=\{128,256,512,1024\}) but the same stride size (\texttt{S}=128).
We observe that BART tends to select content from the first 150 to 200 tokens of the input and add them to the abstract. 
It indicates that the model exhibits strong lead bias even for spoken text, which differs from news writing~\cite{grenander-etal-2019-countering}.
Additionally, we examine the length of BART abstracts, measured by the number of characters in an abstract. 
Using windows from 128 to 1024 tokens, we find that the avg. abstract length increases from 281 to 332 characters, $\approx$56 to 66 words assuming 5 characters per word on average for English texts~\cite{Shannon51}.
While a larger window can lead to a longer abstract, the abstract size is disproportionate to the window size. 
These results are obtained on the training set using human transcripts as input.

\begin{figure}[t]
\centering
\includegraphics[width=2.7in]{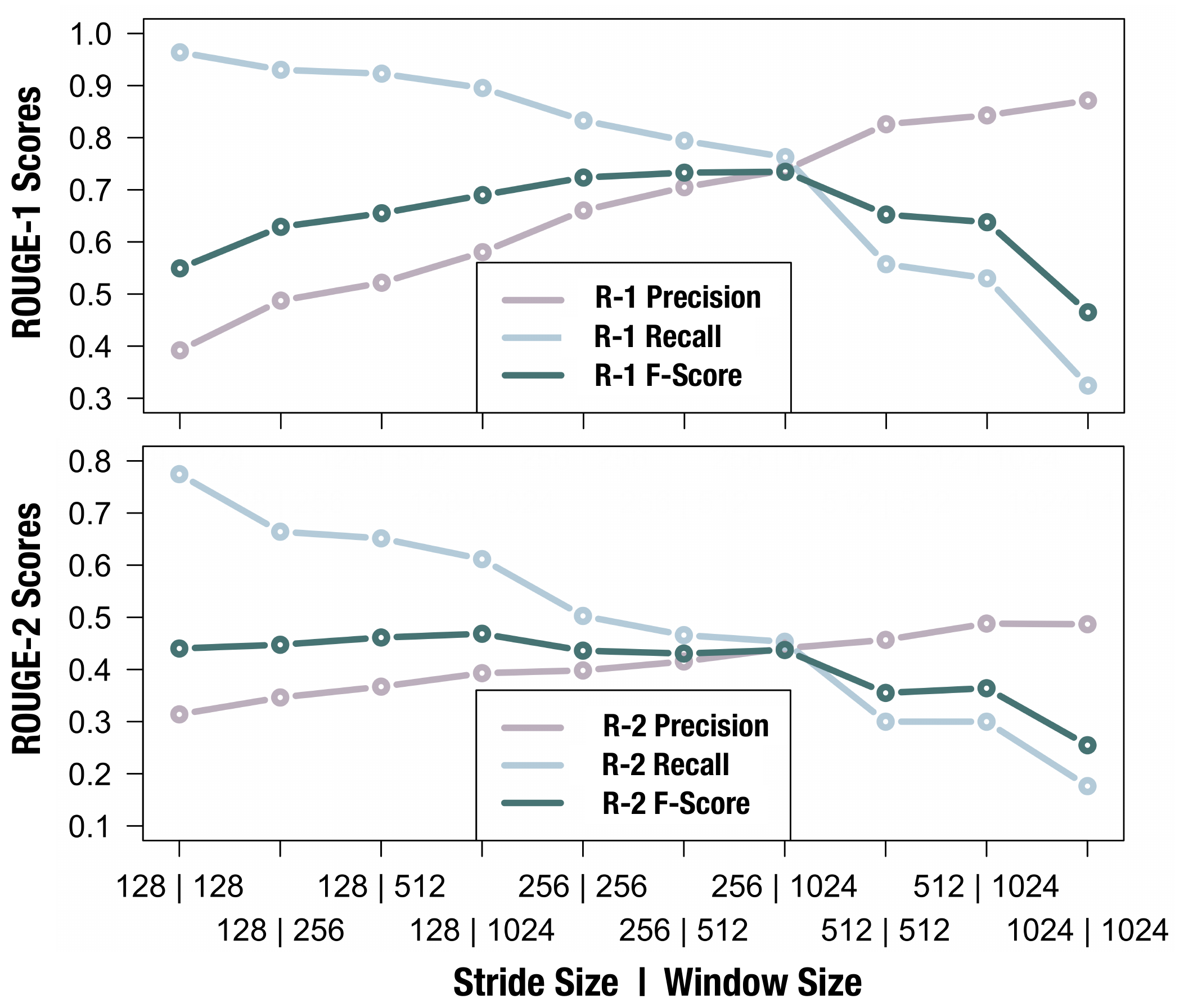}
\vspace{-0.05in}
\caption{R-1 and R-2 scores when different combinations of stride (\texttt{S}) and window (\texttt{W}) sizes are used.
Results are obtained on the ICSI training set for human transcripts.
With (\texttt{S}=256, \texttt{W}=1024), we obtain balanced precision and recall scores.
The best R-2 F-score is achieved with (\texttt{S}=128, \texttt{W}=1024).
}
\label{fig:rouge_train}
\vspace{-0.1in}
\end{figure}

\begin{figure}[t]
\centering
\includegraphics[width=2.9in]{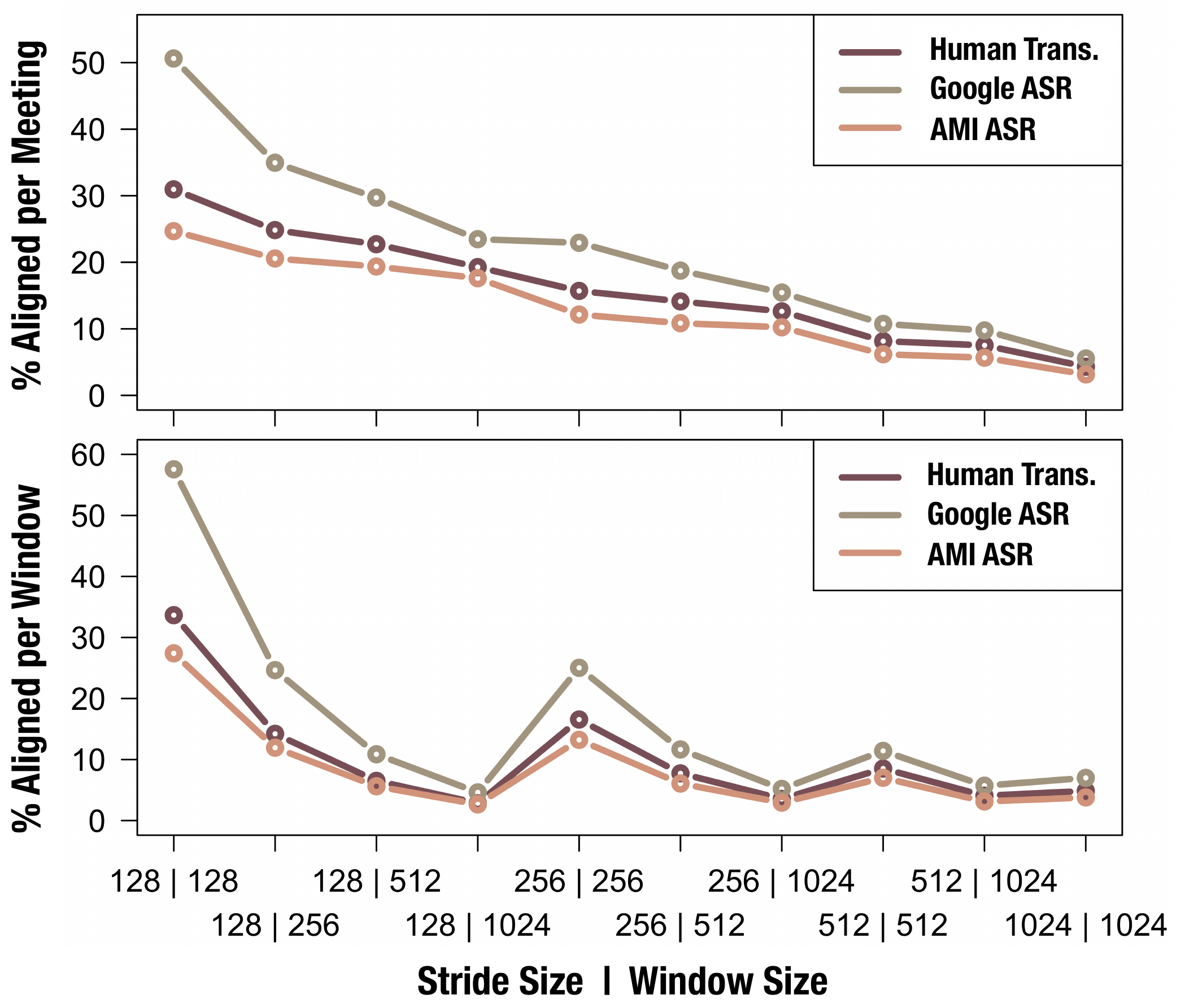}
\vspace{-0.05in}
\caption{
Percentage of supporting utterances per meeting (\textsc{Top}) and per local window (\textsc{Bottom}).
Results are obtained on the ICSI training set with different combinations of stride (\texttt{S}) and window (\texttt{W}) sizes, for human transcripts and two versions of automatic transcripts (Google vs. AMI).
}
\label{fig:alignment}
\vspace{-0.1in}
\end{figure}

\begin{table}[t]
\setlength{\tabcolsep}{4pt}
\renewcommand{\arraystretch}{1.1}
\centering
\begin{fontppl}
\begin{footnotesize}
\begin{tabular}{|l|rrr|}
\hline
 & \multicolumn{3}{c|}{\textbf{Utterance Rating}} \\
\textbf{System} & {Score-2} & {Score-1} & {Score-0} \\
\hline
\hline
TextRank & 8.58\% & 25.66\% & 65.77\% \\
Supervised-BERT & 11.35\% & 28.96\% & 59.69\% \\
\textbf{Sliding Window} & \textbf{11.46}\% & \textbf{26.11}\% & \textbf{62.43}\% \\
\hline
\end{tabular}
\end{footnotesize}
\end{fontppl}
\vspace{-0.05in}
\caption{Percentage of summary utterances rated as highly relevant (2), relevant (1) and irrelevant (0) by human evaluators.
The systems for comparison are TextRank, a supervised BERT summarizer~\cite{koay-etal-2020-domain} and Sliding Window.}
\label{tab:human_eval}
\vspace{-0.1in}
\end{table}

\begin{table*}[t]
\setlength{\tabcolsep}{5pt}
\renewcommand{\arraystretch}{1.1}
\centering
\begin{scriptsize}
\begin{fontppl}
\begin{tabular}{|l|l|lll|}
\hline
\textbf{Speaker} & \textbf{Utterance} & \textbf{BERT} & \textbf{SW} & \textbf{Gold} \\
\hline
\hline
fn002 & I - Hynek last week say that if I have time I can to begin to - to study & {\cellcolor[gray]{.9}}1 & {\cellcolor[gray]{.9}}1 & {\cellcolor[gray]{.9}}1\\
fn002 & well seriously the France Telecom proposal to look at the code and something like that & {\cellcolor[gray]{.9}}1 & {\cellcolor[gray]{.9}}1 & {\cellcolor[gray]{.9}}1\\
me013 & Mm-hmm. & 0 & 0 & 0\\
fn002 & to know exactly what they are doing because maybe that we can have some ideas & {\cellcolor[gray]{.9}}1 & 0 & 0\\
me013 & Mm-hmm. & 0 & 0 & 0\\
fn002 & but not only to read the proposal. Look look & 0 & 0 & 0\\
fn002 & carefully what they are doing with the program and I begin to - to work also in that. & {\cellcolor[gray]{.9}}1 & 0 & {\cellcolor[gray]{.9}}1\\
fn002 & But the first thing that I don't understand is that they & 0 & {\cellcolor[gray]{.9}}1 & {\cellcolor[gray]{.9}}1\\
fn002 & are using & 0 & 0 & {\cellcolor[gray]{.9}}1\\
fn002 & the uh log energy that this quite - I don't know why they have some & 0 & {\cellcolor[gray]{.9}}1 & {\cellcolor[gray]{.9}}1\\
fn002 & constant in the expression of the lower energy. I don't know what that means. & 0 & {\cellcolor[gray]{.9}}1 & {\cellcolor[gray]{.9}}1\\
me018 & They have a constant in there, you said? & 0 & {\cellcolor[gray]{.9}}1 & 0\\
\hline
\end{tabular}
\end{fontppl}
\end{scriptsize}
\vspace{-0.05in}
\caption{Extractive summaries produced by the sliding-window approach (\textsc{SW}) appear to read more coherently than those of the supervised BERT summarizer. 
Consecutive sentences in SW summaries are more likely to be associated with the same idea/speaker compared to supervised-BERT. 
``Gold'' are ground-truth summary utterances.}
\label{tab:example_output}
\vspace{-0.1in}
\end{table*}

In Figure~\ref{fig:results_cls_train}, we investigate various combinations of stride (\texttt{S}) and window sizes (\texttt{W}) and report their precision, recall and F-scores on summary utterance selection.
Similarly, the results are obtained on the training set using human transcripts as input.
We highlight some interesting findings.
We observe that a large context window (\texttt{W}=1024) tends to give high precision. 
A small window combined with small stride yields high recall due to more utterances are selected for the summary.
For example, both settings (\texttt{W}=512, \texttt{S}=128) and (\texttt{W}=1024, \texttt{S}=256) allow an utterance to be visited 4 times.
The former achieves a higher recall (0.395 vs. 0.239) due to its smaller window and stride sizes.
In Figure~\ref{fig:rouge_train}, we show R-1 and R-2 scores obtained on the training set for all combinations of stride and window sizes.
We find that recall scores decrease substantially using large stride sizes (>=512 tokens). 
With (\texttt{S}=256, \texttt{W}=1024), we obtain balanced precision and recall scores.
The best R-2 F-score is achieved with (\texttt{S}=128, \texttt{W}=1024) which is used at test time.

In Figure~\ref{fig:alignment}, we present the percentage of supporting (summary) utterances per meeting and per window, for various combinations of window and stride sizes. 
On human transcripts, we observe that combining small stride and window sizes (\texttt{S}=128, \texttt{W}=128) has led to $\sim$30\% utterances to be selected per meeting. 
In contrast, (\texttt{S}=128, \texttt{W}=1024) selects 19\% of the utterances.
Human transcripts and automatic transcripts generated by AMI ASR appear to show similar behavior, but the Google transcriber breaks up utterances differently.

We further conduct a human evaluation on the six test meetings.
Three human evaluators (two native speakers and a non-native speaker) are employed for this task.
They rate each summary utterance as highly relevant (2), relevant (1) or irrelevant (0) by matching the utterance with the meeting abstract provided by the ICSI corpus.
The systems for comparison are SW, TextRank and the fully supervised BERT summarizer~\cite{koay-etal-2020-domain}.
In Table~\ref{tab:human_eval}, we report the percentage of summary utterances assigned to each category (Fleiss' Kappa=0.29).
Our summarizer obtains promising results.
It outperforms TextRank and performs comparably to supervised-BERT. 
We find that the SW summarizer navigates through the transcript in an \emph{equally detailed} manner.
It leads to coherent and sometimes verbose summaries, compared to other extractive summaries.
A snippet of the transcript and its accompanying summaries are shown in Table~\ref{tab:example_output}.

\section{Conclusion}

We investigate the feasibility of a sliding-window approach to generating meeting minutes and obtain promising results on both human and automatic transcripts.
The approach does not require annotated data and it has a great potential to be extended to meetings of various domains.
Our future work includes, in the near horizon, experimenting with a look-ahead mechanism to enable the summarizer to skip over insignificant transcript segments.

\section*{Acknowledgements}
We thank the anonymous reviewers for their helpful feedback.
We would also like to thank Kaiqiang Song for helping transcribe the meetings using Google's Speech-to-Text API. 
Xiaojin Dai was supported by NSF DUE-1643835.
We thank Amazon for partially sponsoring the research and computation in this study through the Amazon AWS Machine Learning Research Award.

\bibliography{summ,fei,more,anthology}
\bibliographystyle{acl_natbib}

\end{document}